\documentclass[runningheads]{llncs}

\usepackage{eccv}

\usepackage{eccvabbrv}

\usepackage{graphicx}
\usepackage{booktabs}
\usepackage{wrapfig}
\usepackage[accsupp]{axessibility}  

\usepackage[pagebackref,breaklinks,colorlinks,citecolor=eccvblue]{hyperref}
\usepackage{hyperref}

\usepackage{orcidlink}

\usepackage{graphicx} 
\usepackage{hyperref}
\usepackage{parskip}
\usepackage{paralist}
\usepackage{amsmath}
\usepackage{comment}
\usepackage{amssymb}
\usepackage{subcaption}
\usepackage{multirow}
\usepackage{booktabs}
\usepackage{changepage}
\usepackage{hyperref}
\usepackage{url}
\usepackage{xcolor}

\newcommand\greenval{\textcolor{Green}}
\newcommand\redval{\textcolor{Red}}

\newcommand\blfootnote[1]{%
  \begin{NoHyper}%
  \renewcommand\thefootnote{}\footnote{#1}%
  \addtocounter{footnote}{-1}%
  \end{NoHyper}%
}

\graphicspath{{./images/}}

\title{m2mKD: Module-to-Module Knowledge Distillation for Modular Transformers}

\author{
\inst{1}Ka Man Lo$^{\ast}$ \and
\inst{2}Yiming Liang$^{\ast}$\and
\inst{3}Wenyu Du \and
\inst{4}Yuantao Fan \and \\
\inst{5}Zili Wang \and
\inst{6}Wenhao Huang \and
\inst{7}Lei Ma$^{\dag}$\and
\inst{8}Jie Fu$^{\dag}$
}

\authorrunning{KM.~Lo et al.}

\institute{$^{1}$ University of Macau $^{2}$ Institute of Automation, Chinese Academy of Sciences \\
$^{3}$ University of Hong Kong $^{4}$ Beijing University of Posts and Telecommunications \\
$^{5}$ Independent $^{6}$ 01.AI $^{7}$ Peking University $^{8}$ HKUST 
}

\begin{document}

\titlerunning{m2mKD}
\maketitle

\begin{abstract}
Modular neural architectures are gaining attention for their powerful generalization and efficient adaptation to new domains. 
However, training these models poses challenges due to optimization difficulties arising from intrinsic sparse connectivity. 
Leveraging knowledge from monolithic models through techniques like knowledge distillation can facilitate training and enable integration of diverse knowledge.
Nevertheless, conventional knowledge distillation approaches are not tailored to modular models and struggle with unique architectures and enormous parameter counts.
Motivated by these challenges, we propose module-to-module knowledge distillation (\texttt{m2mKD}) for transferring knowledge between modules. 
\texttt{m2mKD} combines teacher modules of a pretrained monolithic model and student modules of a modular model with a shared meta model respectively to encourage the student module to mimic the behaviour of the teacher module.
We evaluate \texttt{m2mKD} on two modular neural architectures: Neural Attentive Circuits (NACs) and Vision Mixture-of-Experts (V-MoE).
Applying \texttt{m2mKD} to NACs yields significant improvements in IID accuracy on Tiny-ImageNet (up to 5.6\%) and OOD robustness on Tiny-ImageNet-R (up to 4.2\%).
Additionally, the V-MoE-Base model trained with \texttt{m2mKD} achieves 3.5\% higher accuracy than end-to-end training on ImageNet-1k. 
Code is available at \url{https://github.com/kamanphoebe/m2mKD}.
\end{abstract}

\blfootnote{$^{\ast}$ Equal contribution.}
\blfootnote{$^{\dag}$ Corresponding authors.}

\section{Introduction}

Despite the success of large monolithic models in various domains, concerns have emerged regarding their limited generalization ability and increasing computational costs. 
Meanwhile, modular models have gained attention as promising alternatives to mitigate these drawbacks.
In contrast to monolithic models with fixed computational graphs and parameters, modular neural architectures dynamically adapt their parameters to the input, offering favorable properties that are absent in static monolithic models \cite{han2021dynamic}.  
Unlike monolithic models that optimize parameters collectively, modular models consist of independent modules that can be updated locally without affecting other network parts. 
These modules specialize in specific tasks and improve generalization performance by activating relevant modules for each input, even for out-of-distribution (OOD) samples \cite{pfeiffer2023modular}.
For instance, DEMix Layers \cite{gururangan2021demix} jointly represent COVID-19-related data using medical and news modules.
Moreover, modular models enhance computational efficiency through conditional calculation.
Mixture-of-Experts (MoE), a typical modular architecture \cite{shazeer2017outrageously}, significantly increases model capacity while maintaining similar computational requirements to the original model \cite{jia2021scaling, lepikhin2020gshard}.


Although modular architectures surpass monolithic models in terms of OOD robustness and computational efficiency, there is still vast room for improvement in their algorithms and implementations. 
Training modular models presents challenges due to optimization difficulties arising from sparse interactions. 
While recent works \cite{zoph2022st, mustafa2022multimodal, nie2021evomoe} have investigated the training instability of modular models, thorough studies in this area still need to be completed. 
On the engineering side, several implementations of MoE exist, but many of them \cite{he2022fastermoe, lepikhin2020gshard} do not consider dynamic workload allocation for experts or support acceleration techniques like mixed precision training. 
Although adaptive parallelism \cite{hwang2023tutel} has addressed the dynamic nature of MoEs, training MoEs remains considerably slower than their monolithic counterparts due to unavoidable communication overheads. 

To alleviate the optimization issue, employing a pretrained monolithic model to guide the training of modular models shows potential. 
\textbf{Knowledge distillation (KD)} \cite{hinton2015distilling} is an existing technique that transfers knowledge from a pretrained teacher model to a smaller student model. 
KD has proven effective in the context of monolithic models. 
However, directly applying conventional KD approaches to modular models is computationally expensive due to their large model sizes. 
Using a monolithic model as the teacher for a larger modular model may even harm performance (see \cref{tab:moe-res}). 
Furthermore, monolithic models trained by regular methods may not be the optimal choice for teachers. 

\begin{figure*}[t]
    \centering
    \includegraphics[width=0.75\linewidth]{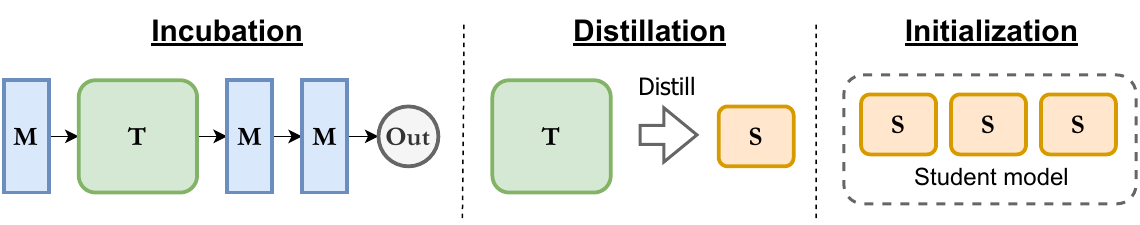}
    \caption{Overview of \texttt{m2mKD}. First, incubate the teacher modules $T$ using a meta model $M$. Then, distill the knowledge from the teacher modules to the student modules. Finally, initialize the modular model with the parameters from the student modules $S$.}
    \label{fig:m2mKD}
\end{figure*}

Inspired by the divide-and-conquer training mechanism of Deep Incubation \cite{ni2023deep}, we introduce \textbf{module-to-module knowledge distillation} (\texttt{m2mKD}) to transfer knowledge between sub-modules of the monolithic teacher and the modular student. 
As illustrated in Figure \cref{fig:m2mKD}, we first adopt Deep Incubation, a modular training method, to incubate the teacher modules using a small meta-model. 
Next, we encourage the student module to imitate the behaviour of the teacher module. 
Finally, the distilled student modules are used to initialize the modular student model. 
By adapting the distillation process at the distributed module level, \texttt{m2mKD} significantly reduces the capacity requirement of the teacher model and enables independent training, making distillation less computationally expensive. 
The cost savings become more noticeable when the modular student is large or when multiple teachers are involved, such as in ensemble learning. 
Moreover, the teacher model is trained in a modular way, which may benefit teaching a modular student. 
Note that our \texttt{m2mKD} algorithm does not impose any restrictions on the architecture of both the teacher and student models. 
We evaluate the performance of \texttt{m2mKD} on both NACs and V-MoE.
Experiments show that a NAC model trained using \texttt{m2mKD} improves IID accuracy by approximately 5\% on Tiny-ImageNet and enhances out-of-distribution (OOD) robustness by about 4\% on Tiny-ImageNet-R. 
Additionally, there is an average gain of approximately 1\% on ImageNet \cite{deng2009imagenet} and ImageNet-R \cite{hendrycks2021many}. 
The experimental results for V-MoE models indicate that \texttt{m2mKD} also works in the case of a small teacher.

Our contributions are as follows:
\begin{inparaenum}[(1)]
    \item To the best of our knowledge, we are the first to investigate knowledge distillation for modular models. 
    \item We demonstrate the challenges associated with distilling modular models and introduce a tailored algorithm to address these challenges. 
    \item Our approach can be seen as a promising framework for transforming a monolithic model into a modular model with arbitrary neural architecture. Notably, this transformation is developed in a modular-style manner.
    \item The proposed method is capable of handling irregular distillation scenarios where the student model size is larger than the teacher model size. It has the potential to work not only for the monolithic-to-modular case but also for the monolithic-to-monolithic case. 
    \item We verify the feasibility of using Deep Incubation for modular models.
\end{inparaenum}

\section{Related Work}

\textbf{Knowledge distillation.} Knowledge distillation is a model compression technique that transfers knowledge from a large teacher network to a smaller student model \cite{hinton2015distilling}. 
Common knowledge distillation approaches involve mimicking the teacher model's softened output \cite{hinton2015distilling, kim2018paraphrasing, ba2014deep}. 
In scenarios with multiple teachers, the outputs of all teacher models can be averaged \cite{hinton2015distilling}, or other methods can be used \cite{fukuda2017efficient, kwon2020adaptive, yuan2021reinforced}. 
Traditional knowledge distillation methods often struggle when a significant capacity gap exists between the student and the teacher models. 
Recent approaches have attempted to address this gap by using teacher assistants \cite{mirzadeh2020improved, son2021densely}. 
In addition, some existing works conduct distillation at the module level instead of distilling entire models \cite{xu2020bert, zhao2022knowledge, liang2024module}. 
While previous works primarily focus on compressing model sizes within the context of monolithic models, we propose a module-to-module distillation technique to enhance the performance of modular architectures, especially in unconventional distillation scenarios.
Particularly, the distillation of modules is mutually independent and can be executed in a distributed manner. 

\textbf{Modularization.} Modular deep learning involves decomposing neural architectures into independent and parameter-efficient modules, where samples are conditionally routed to subsets of modules, and their outputs are aggregated. 
Modularization has been widely applied in various areas such as transfer learning \cite{li2021prefix, houlsby2019parameter, platanios2018contextual, hu2021lora}, modular training \cite{ni2023deep}, and scaling up model size \cite{jia2021scaling, lepikhin2020gshard}.

In transfer learning, pre-trained models are often used as modules in an assembled model for new tasks. 
This assembled model can adapt to new data by adjusting or adding modules to enhance performance in different scenarios. 
For instance, Brown et al. \cite{brown2020language} added a prompt module to the input of a pre-trained model, while Rebuffi et al. \cite{rebuffi2018efficient} introduced adapter modules into the model architecture. 
Modular training approaches like Deep Incubation \cite{ni2023deep} incubate modules in individual nodes to avoid communication overhead and accelerate convergence. Conditional computation enables the scaling of model size while maintaining inference complexity. 
V-MoE \cite{jia2021scaling} scales up vision models with only half the computation required during inference. 
The MoE framework introduces experts to enable modularization at the FFN layer.
Other modular model architectures include NACs \cite{weiss2022neural}, as well as \cite{shen2023moduleformer, goyal2021coordination}.



\textbf{Monolithic to modular.} There are several efforts to convert monolithic models into modular architectures. 
MoEfication \cite{zhang2021moefication} directly splits the FFN layers of a monolithic model into multiple experts to form MoE layers, while Sparse Upcycling \cite{komatsuzaki2022sparse} copies the MLP parameters to the corresponding experts in the MoE layers. 
In contrast to these works, our proposed \texttt{m2mKD} does not make assumptions about the model architecture and can be applied to any model.


\section{Method}

In this section, we first review the essential aspects of Deep Incubation, as our work heavily relies on it. 
Subsequently, we introduce our proposed method. For convenience, we provide a table (Appendix A) that lists the notations commonly used in this paper.

\begin{figure*}[tb]
    \centering
    \includegraphics[width=.7\linewidth]{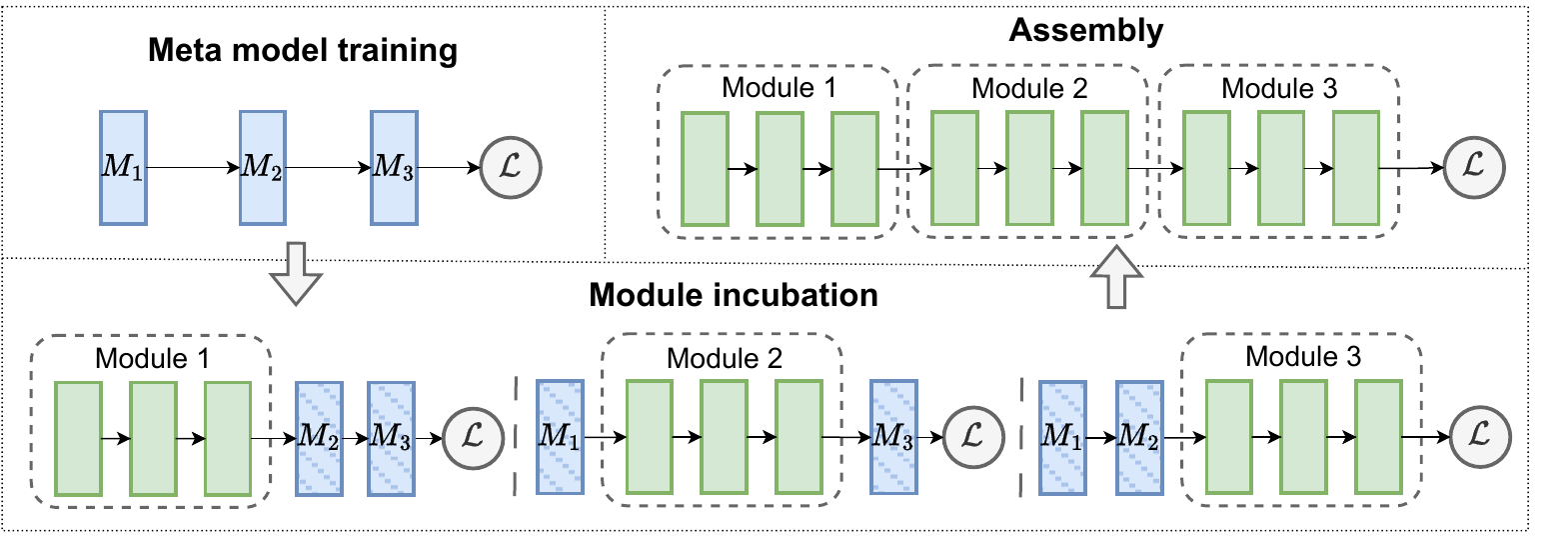}
    \caption{Overview of Deep Incubation. First, a small meta model is pretrained. Then, the module replaces the corresponding layer in the meta model, and only the module's parameters are updated during incubation. Finally, the incubated modules are assembled and fine-tuned to obtain the final model.}
    \label{fig:DeepIncubation}
\end{figure*}

\begin{figure*}[tb]
    \centering
    \includegraphics[width=\linewidth]{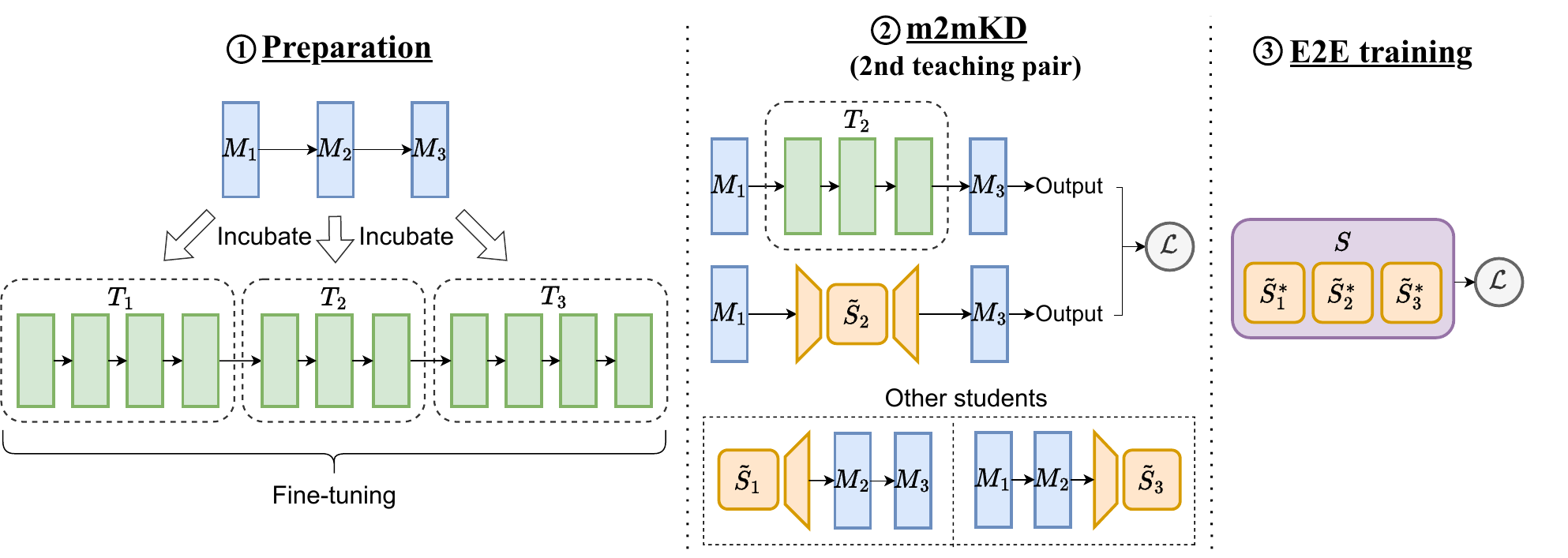}
    \caption{Pipeline of \texttt{m2mKD} ($L=3$). The \texttt{m2mKD} pipeline involves several steps. In the preparation phase, we begin by training a meta model $\mathcal{M}$, incubating the teacher modules $\{T_i\}$, and fine-tuning the assembled teacher model $\mathcal{T}$. Subsequently, the \texttt{m2mKD} process is executed as follows: two hybrid networks are constructed by replacing the $i$-th meta layer $M_i$ with the $i$-th teaching pair $(T_i, \tilde{S}_i)$, and their outputs are compared. Note that except for the stitched student module $\tilde{S}_i$, all other layers are frozen during this phase. Finally, the distilled student modules $\{S_i^*\}$, with stitched layers discarded, are loaded into the student model $\mathcal{S}$ at the beginning of the end-to-end training phase.}
    \label{fig:pipeline}
\end{figure*}

\subsection{Preliminary: Deep Incubation} \label{preliminary}

Deep Incubation \cite{ni2023deep} employs a divide-and-conquer strategy to train large models modularly, as illustrated in \cref{fig:DeepIncubation}. 
The process consists of three stages: meta-model pre-training, module incubation, and assembly. 
In the initial stage, a small model is pre-trained as a meta model using an end-to-end training approach. 
Subsequently, each module replaces the corresponding layer in the meta-model, and only the module parameters are updated during the incubation process. 
The incubation of modules is mutually independent and can be completed in a distributed manner.
Finally, the incubated modules are assembled and fine-tuned to obtain the final model. 
It is worth noting that the number of layers in the meta model corresponds to the number of modules incubated during the incubation stage.

\subsection{Pipeline}


Our method follows a pipeline consisting of three steps (\cref{fig:pipeline}): preparation, module-to-module knowledge distillation (\texttt{m2mKD}), and end-to-end (E2E) training. 
More details are presented below.

\textbf{Preparation.} Prior to commencing the distillation process, the Incubation algorithm is applied to prepare a meta model and a teacher model. 
This step is actually exclusive of the proposed method. 
Assuming that the target student model $\mathcal{S}$ comprises a total of $L$ modules, the meta model $\mathcal{M}$ should have an equivalent number of meta layers.
Similarly, the teacher model $\mathcal{T}$, which consists of $n$ layers ($n \geq L$), is divided into $L$ sub-modules: 
\begin{equation}
    \begin{split}
        \mathcal{S} & = S_L \circ S_{L-1} \circ ... \circ S_{1} \\
        \mathcal{M} & = M_L \circ M_{L-1} \circ ... \circ M_{1} \\
        \mathcal{T} & = T_L \circ T_{L-1} \circ ... \circ T_{1}.
    \end{split}\label{equation 1}
\end{equation}
As proposed by \cite{ni2023deep}, the initial step involves training the meta model in an end-to-end fashion to incubate the $L$ sub-modules $T_1, T_2, ..., T_L$. 
Subsequently, these resulting sub-modules are reassembled to form the teacher model $\mathcal{T}$, which undergoes fine-tuning. 
While we follow the same pipeline described in the original paper, our focus is on the sub-modules themselves rather than the entire model. 
We opt for the Incubation approach instead of merely separating a pre-trained model because we claim that the module incubation phase imparts additional knowledge to the sub-modules. 
This allows them to learn how to function as individual modules rather than being incomplete fragments of a whole model. 

\textbf{Module-to-module knowledge distillation.} Once the fine-tuning of the assembled model $\mathcal{T}$ is complete, the sub-modules, or we call ``teacher modules" hereafter, are ready for running \texttt{m2mKD}. 
Unlike conventional knowledge distillation approaches, \texttt{m2mKD} aims to transfer knowledge between modules rather than entire models. 
Similar to the module incubation proposed by \cite{ni2023deep}, we separately link the teacher and student module to the meta model.
By comparing the outputs of the two resulting hybrid models, the student modules are encouraged to mimic the behaviour of the corresponding teacher modules. 
Previous research by Yang et al. \cite{yang2022deep} demonstrates that blocks located at similar depths in different networks can be considered functionally equivalent. 
This insight suggests that neural networks learn similar patterns at similar network stages. 
Exploiting this insight, we assign teacher modules to students at the same depth, resulting in $L$ teaching pairs $(T_i, \tilde{S}i)|{i=1,...,L}$. 
Each teaching pair can then be performed \texttt{m2mKD} in parallel.
For the $i$-th teaching pair, we replace the $i$-th meta layer with $T_i$ and the stitched student module $\tilde{S}_i$, giving rise to two hybrid networks: 
\begin{equation}
    \begin{split}
        \tilde{\mathcal{M}}^{(i)}_T & =  M_L \circ \ldots \circ M_{i+1} \circ T_i \circ M_{i-1} \circ \ldots \circ M_1 \\
        \tilde{\mathcal{M}}^{(i)}_{\tilde{S}} & =  M_L \circ \ldots \circ M_{i+1} \circ \tilde{S}_i \circ M_{i-1} \circ \ldots \circ M_1.
    \end{split}
\end{equation}
The modified student module, denoted as $\tilde{S}_i$, incorporates a linear stitch layer that is inserted right before and/or after the module. 
This stitch layer is responsible for adjusting the dimension of the feature vectors to address any potential dimension mismatch between the meta layer and the student module. 
The weight matrix of the pre-stitch layer is denoted as $W \in \mathbb{R}^{d_M\times d_S}$, while the post-stitch layer has a weight matrix denoted as $W \in \mathbb{R}^{d_S\times d_M}$. 
Here, $d_M$ represents the dimension of the meta layer, and $d_S$ represents the dimension of the student module.

The two hybrid networks are now considered as ``complete" models. 
Consequently, we can directly apply the conventional response-based knowledge distillation technique. 
Specifically, for the classification problems considered in this paper, we compare the output logits $z_T$ and $z_{\tilde{S}}$ given an input $x$ by measuring the Kullback-Leibler (KL) divergence {hinton2015distilling}. 
The total loss $\mathcal{L}$ is then defined as the weighted sum of the classification cross-entropy loss $L_{CE}$ and the knowledge distillation loss $L_{KD}$: 
\begin{equation}
    \label{Eq:loss}
    \begin{split}
        \mathcal{L} & = L_{CE} + \alpha L_{KD} \\
        L_{CE} & = H(softmax(z_{\tilde{S}}), y) \\ 
        L_{KD} & = \tau^2KL(softmax(z_{\tilde{S}}/\tau), softmax(z_T/\tau)) 
    \end{split}
\end{equation}
where $\alpha$ represents the balancing factor, $\tau$ denotes the softmax temperature, and $y$ stands for the label associated with the input $x$. 
Throughout the distillation process, only the student module is updated, while both the meta layers and the teacher module remain frozen.

\textbf{End-to-end training.} Given a student model $\mathcal{S}$ consisting of $L$ modules, we can run \texttt{m2mKD} for all teaching pairs $(T_i, \tilde{S}i)|{i=1,...,L}$ in parallel to obtain $L$ distilled student modules $\tilde{S}_i^*$. 
The last step is to simply load the learned parameters into $\mathcal{S}$ and perform end-to-end (E2E) training. 
Note that all stitch layers of the student modules are discarded at this stage. 
For the NAC model architecture, there can be a dimension mismatch problem when loading the learned parameters for certain components, especially if the datasets used in \texttt{m2mKD} and E2E training differ (refer to \cref{sec:nac-res}). 
In such cases, these incompatible elements will not be loaded.

\section{Experiments on NACs}

\begin{figure*}[tb]
    \centering
    \begin{subfigure}[b]{0.53\linewidth}
        \includegraphics[width=1\linewidth]{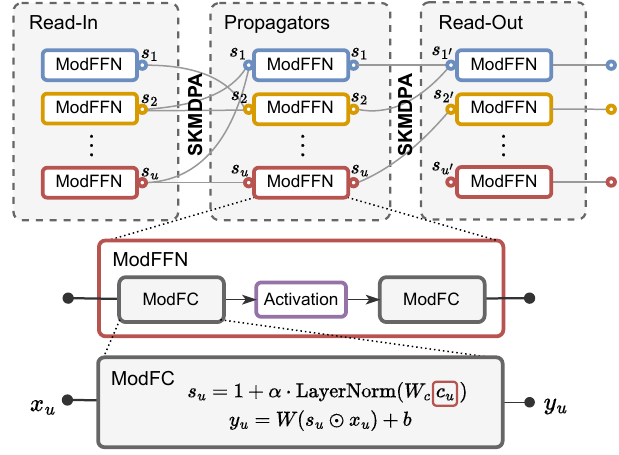}
        \caption{NAC model}
        \label{fig:NACs}
    \end{subfigure}
    \begin{subfigure}[b]{0.43\linewidth}
        \includegraphics[width=1\linewidth]{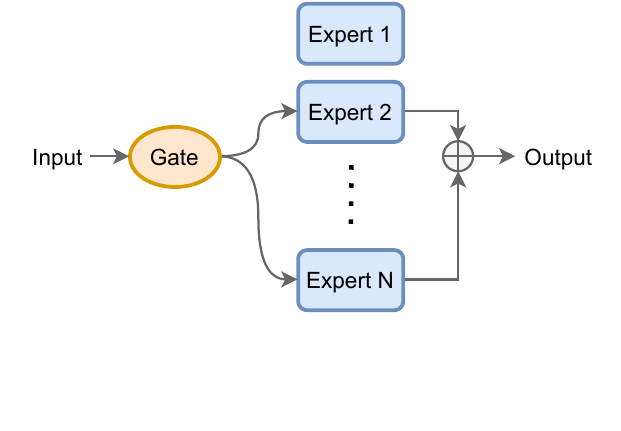}
        \caption{MoE layer}
        \label{fig:MoE}
    \end{subfigure} 
    \caption{Overview of the two model architectures used in the experiments. (a) NAC consists of a read-in layer, multiple propagator layers, and a read-out layer. Each layer of modules is sparsely connected through SKMDPA. The ModFC layer serves as the fundamental computing unit within each module and forms ModFFNs. (b) The MoE layer is constructed using multiple experts and a gate, which selects a specified number of experts (in this case, 2) to process the inputs. The outputs of the selected experts are subsequently aggregated.}
\end{figure*}


NAC is a novel modular architecture as depicted in \cref{fig:NACs}. 
It comprises a read-in layer, multiple propagator layers, and a read-out layer. 
Modules within two consecutive layers are connected sparsely using Stochastic Kernel Modulated Dot-Product Attention (SKMDPA). 
The basic computing unit within each module is the ModFC layer, which forms ModFFNs. 
The majority of parameters are shared across modules, and each module conditions its computation using its own code vector $c_u$. 
The SKMDPA mechanism employs sparse attention to calculate the similarity between modules based on their signature vector $s_u$, thereby determining the communication between modules. 
Each module maintains a state vector $\theta_u$, which serves as the input for the subsequent layer. 
The state vector is initialized in the read-in layer, updated through multiple propagator layers, and eventually used as the input for the read-out layer. 

\subsection{Setups}

\textbf{Datasets.} In this paper, we focus on the image classification task. The training for both the preparation and \texttt{m2mKD} phases is conducted using the ImageNet-1k dataset \cite{deng2009imagenet}. 
For the NAC models, we perform end-to-end training using both the ImageNet dataset and its miniature version, Tiny-ImageNet. 
For OOD evaluation, we utilize the ImageNet-R(enditions) dataset \cite{hendrycks2021many} and its down-sampled subset, Tiny-ImageNet-R. 
Additionally, the CIFAR-100 dataset \cite{krizhevsky2009learning} is employed for few-shot adaptation.

\textbf{Preparation.} Since our target NAC model consists of a total of 10 layers ($L=10$), we choose the DeiT-Huge model \cite{dosovitskiy2020image} with 32 layers as the teacher model to ensure sufficient depth. 
The DeiT-Huge model is divided into 10 sub-modules, with the first and last sub-modules containing 4 layers each, and the remaining sub-modules comprising 3 layers each. 
The NAC models contain no more than 37M parameters, while the DeiT-Huge model contains 632M parameters. 
This yields approximately 3.7M and 63.2M parameters for each student and teacher module, respectively. 
At the beginning of the preparation stage, a 10-layer meta model is trained on ImageNet for 300 epochs. 
Subsequently, each teacher module is incubated by the meta model for 100 epochs, and all modules are assembled back together for additional fine-tuning of 100 epochs. 
The training configurations for this stage are identical to the Deep Incubation approach.

\textbf{m2mKD.} When the teacher part is hidden, \texttt{m2mKD} can be seen as module incubation with an additional loss term. 
Therefore, we again adopt similar experimental settings from module incubation for \texttt{m2mKD}. 
The only difference is that each student module is trained for just 10 epochs. 
We set the balancing factor $\alpha$ to 0.5 and the softmax temperature $\tau$ to 1.0. 
Given the dimension of the meta model $d_M=1280$ and of the student modules $d_S=384$ for Tiny-ImageNet\footnote{The dataset here refers to the one used in the E2E training phase.} (or $d_S=512$ for ImageNet), there would be 1M parameters for each pair of stitch layers. 
After connecting the student and teacher modules with meta layers, the resulting hybrid networks $\tilde{M}_S^{(i)}$ and $\tilde{M}_T^{(i)}$ contain 182.5M and 242.0M parameters, respectively. 

Originally, the NAC propagator layers receive the state vectors of the previous layer as inputs and update the states after the computation of SKMDPA and ModFFN. 
However, we remove the SKMDPA part for our student modules, allowing all modules to communicate with each other. 
This modification is made for two reasons: 
\begin{inparaenum}[(1)]
    \item The signature vectors which determine the communication probability between modules are shared across depths. Even though we learn a set of signature vectors for each student module, they cannot be reused during the end-to-end training of NAC models.
    \item The inputs of the student modules are changed to be the feature vectors from the meta layers and  no longer represent the states of modules. These inputs contain all the necessary information, and none of them should be omitted. 
\end{inparaenum}
To ensure that the ModFC layers function properly, we randomly initialize the state and code vectors. 
Note that these vectors are neither updated during the distillation process nor loaded into the target model for the end-to-end training. 
Specifically, $\tilde{S}_1$ consists of the input tokenizer and the read-in layer, $\tilde{S}_{i=2,...,L-1}$ consists of the $i$-th modified propagator layer, and $\tilde{S}_L$ includes both the modified read-out layer and the output tokenizer. 

\textbf{E2E training.} When training on the same dataset, the hyperparameters remain unchanged for both the reproduced baselines and the distilled NAC models. 
Since the original paper on NACs does not provide a comprehensive list of hyperparameters and we were unable to reproduce the reported results using the given hyperparameters, we adjusted some of them in order to approach their reported results as closely as possible. 
Appendix B presents a selection of our hyperparameters, including all altered values.

\subsection{Results} \label{sec:nac-res}

\begin{table*}[tb]
  \footnotesize
  \centering
  \caption{IID and OOD performance of NACs trained on Tiny-ImageNet. Our \texttt{m2mKD} approach is compared against the original end-to-end training.}
  \label{tab:tiny-result}
  \begin{tabular}{llllll}
    \toprule
    && \multicolumn{2}{l}{IID} & \multicolumn{2}{l}{OOD} \\
    && \multicolumn{2}{l}{(Tiny-ImageNet)} & \multicolumn{2}{l}{(Tiny-ImageNet-R)} \\
    \midrule
    & Graph prior & Acc@1 & Acc@5 & Acc@1 & Acc@5 \\
    \midrule
    \multirow{4}{*}{NACs} & Scale-Free & 60.83 & 82.35 & 20.74 & 41.71 \\
                          & Planted-Partition & 60.57 & 82.20 & 20.50 & 42.94 \\
                          & Ring-of-Cliques & 60.70 & 82.57 & 20.89 & 41.80 \\
                          & Erdos-Renyi & 61.53 & 83.14 & 21.08 & 42.25 \\
    \hline
    \multirow{4}{*}{$+\texttt{m2mKD}$\ } & Scale-Free & 66.47 \greenval{$\uparrow$5.64} & 85.08 \greenval{$\uparrow$2.73} & 24.31 \greenval{$\uparrow$3.57} &  44.38 \greenval{$\uparrow$2.67}\\
                          & Planted-Partition & 66.04 \greenval{$\uparrow$5.47}& 85.68 \greenval{$\uparrow$3.48}& 24.69 \greenval{$\uparrow$4.19}& 45.36 \greenval{$\uparrow$2.42}\\
                          & Ring-of-Cliques & 66.21 \greenval{$\uparrow$5.51}& 85.49 \greenval{$\uparrow$2.92}& 24.89 \greenval{$\uparrow$4.00}& 45.74 \greenval{$\uparrow$3.94}\\
                          & Erdos-Renyi & 65.99 \greenval{$\uparrow$4.46}& 85.37 \greenval{$\uparrow$2.23}& 24.54 \greenval{$\uparrow$3.46}& 45.45 \greenval{$\uparrow$3.20}\\
    \bottomrule
  \end{tabular}
\end{table*}

\begin{table*}[tb]
  \footnotesize
  \centering
  \caption{IID and OOD performance of NACs trained on ImageNet. Our \texttt{m2mKD} approach is compared against the original end-to-end training.}
  \label{tab:nacs-imgnet-result}
  \begin{tabular}{llllll}
    \toprule
    && \multicolumn{2}{l}{IID} & \multicolumn{2}{l}{OOD} \\
    && \multicolumn{2}{l}{(ImageNet)} & \multicolumn{2}{l}{(ImageNet-R)} \\
    \midrule
    & Graph prior & Acc@1 & Acc@5 & Acc@1 & Acc@5 \\
    \midrule
    \multirow{4}{*}{NACs} & Scale-Free & 75.61 & 93.93 & 37.30 & 54.01 \\
                          & Planted-Partition & 75.71 & 94.09 & 37.63 & 54.38 \\
                          & Ring-of-Cliques & 76.12 & 94.35 & 37.21 & 53.70 \\
                          & Erdos-Renyi & 75.71 & 93.95 & 36.48 & 53.39 \\
    \hline
    \multirow{4}{*}{$+\texttt{m2mKD}\ $} & Scale-Free & 76.63 \greenval{$\uparrow$1.02} & 94.62 \greenval{$\uparrow$0.69} & 39.18 \greenval{$\uparrow$1.88} & 55.10 \greenval{$\uparrow$1.09} \\
                          & Planted-Partition & 76.49 \greenval{$\uparrow$0.78} & 94.01 \redval{$\downarrow$0.08} & 38.02 \greenval{$\uparrow$0.39} & 53.58 \redval{$\downarrow$0.80} \\
                          & Ring-of-Cliques & 76.89 \greenval{$\uparrow$0.77} & 94.44 \greenval{$\uparrow$0.09} & 39.29 \greenval{$\uparrow$2.08} & 55.42 \greenval{$\uparrow$1.72} \\
                          & Erdos-Renyi & 76.56 \greenval{$\uparrow$0.85} & 94.39 \greenval{$\uparrow$0.44} & 38.84 \greenval{$\uparrow$2.36} & 54.69 \greenval{$\uparrow$1.30} \\
    \bottomrule
  \end{tabular}
\end{table*}

\textbf{Main results.} As aforementioned, our teacher model DeiT-Huge and NAC student modules in the first two phases are trained solely on ImageNet. 
The assembled teacher model achieves a validation accuracy of 81.8\%. 
With 8 A100 80GB GPUs, the training time for all ten student modules in the \texttt{m2mKD} phase is under 12 hours. 
For E2E training on Tiny-ImageNet, we need to discard a portion of the input tokenizer and the entire output tokenizer in the student modules due to dimension mismatch.
\cref{tab:tiny-result} compares the reproduced baselines and our distilled NAC models on Tiny-ImageNet and Tiny-ImageNet-R. 
The reproduced results slightly outperform the reported values in the original paper. 
Although the dataset is different from the one used in the \texttt{m2mKD} phase, our distilled models with various graph prior regularizers exhibit an average improvement of 5.3\% in IID performance and 3.8\% in OOD robustness over the baselines. 
This indicates that the addition of distilled student modules not only enhances the ability of the final NAC for similar tasks, but also improves its modularity. 
To ensure reproducibility, we repeat the E2E training phase of the distilled model with scale-free prior three times.
The results are presented in Appendix C.

The comparison for ImageNet is shown in \cref{tab:nacs-imgnet-result}. 
The original paper reports a validation accuracy of 77\% for the NAC trained on ImageNet with a scale-free graph prior, while we reproduce the baselines for all four graph priors, achieving a maximum value of 76.1\%.
Our distilled NACs achieve maximum gains of 1.0\% and 2.4\% for IID and OOD performance, respectively.

\begin{figure}[thbp]
    \begin{minipage}[t]{0.32\textwidth}
        \centering
        \includegraphics[width=\textwidth]{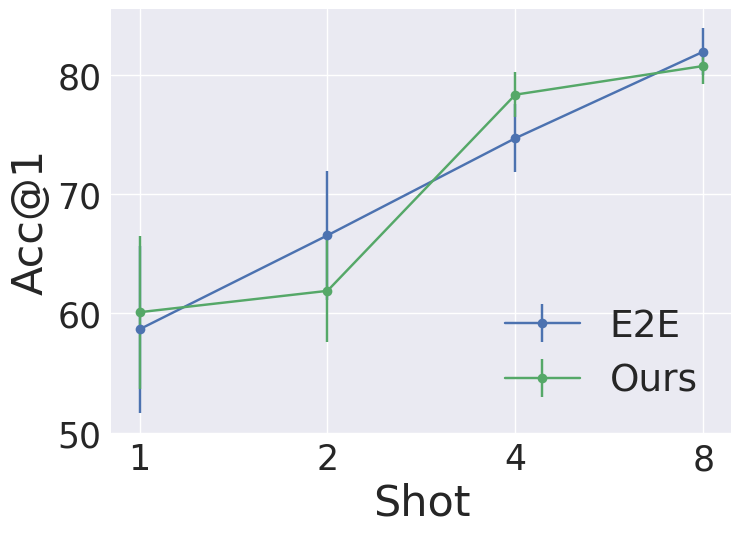}
        \caption{Few-shot performance of NACs on CIFAR-100.}
        \label{fig:nac-fewshot}
    \end{minipage}
    \begin{minipage}[t]{0.01\textwidth}
        ~~
    \end{minipage}
    \begin{minipage}[t]{0.65\textwidth}
        \centering
       \begin{subfigure}{.49\textwidth}
            \includegraphics[width=\textwidth]{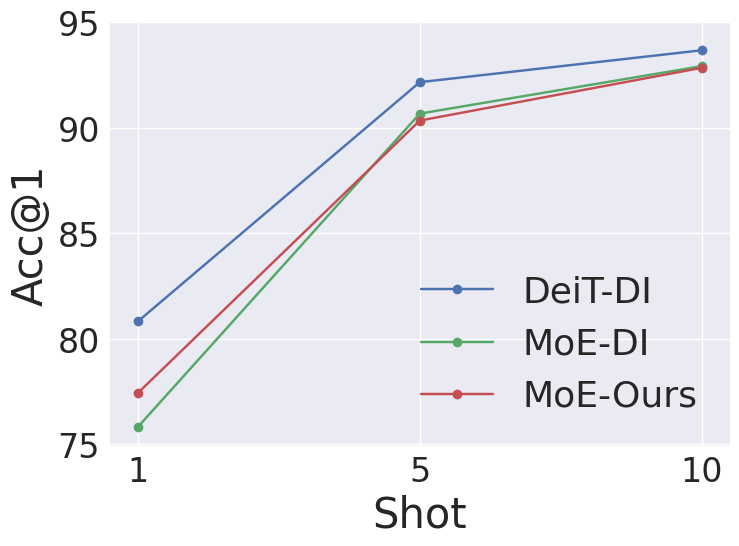}
        \end{subfigure}
        \begin{subfigure}{.49\textwidth}
            \includegraphics[width=\textwidth]{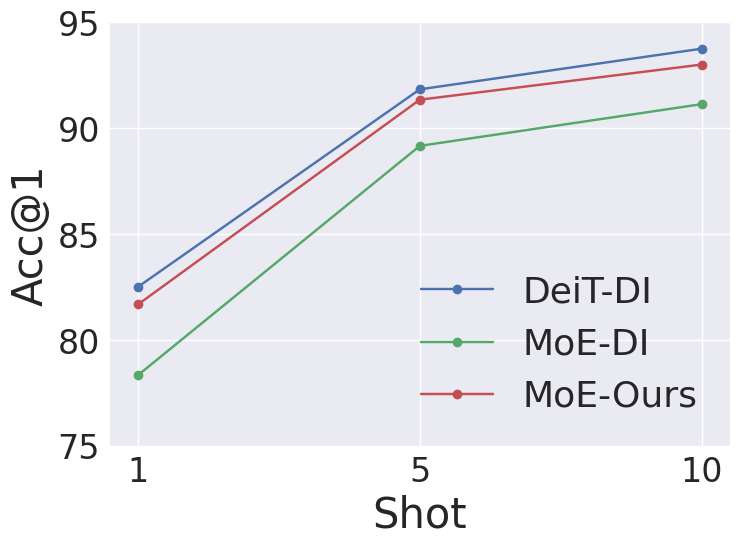}
        \end{subfigure}
        \caption{Few-shot performance of DeiT and V-MoE-B models on CIFAR-100 (left) and CUB-100 (right). DI denotes Deep Incubation.}
        \label{fig:moe-fewshot}
    \end{minipage}
\end{figure}

\textbf{Few-shot.} To evaluate the few-shot adaptation performance, we further fine-tune the classifier layer of the distilled NAC with a scale-free prior, which is trained on ImageNet, using a small number of samples from the CIFAR-100 dataset. 
The hyperparameters and a comprehensive table of results can be found in Appendix B and Appendix D, respectively. 
We conduct the experiments with 5 different seeds and report the averaged accuracies and corresponding standard deviations in \cref{fig:nac-fewshot}. 
Our reproduced baselines are at most 10\% higher than the original results. 
It can be observed that the distilled model performs similarly to the baseline with no significant improvement. 
The standard deviations are relatively large, possibly due to the limited number of repetitions (i.e., the number of seeds). 
To examine the variation under the same seeds, we rerun the 2-shot experiments once for all five seeds and find that the largest difference reaches around 7\% given a fixed seed (see Appendix D for details). 
Therefore, additional experiments may be necessary to further validate the few-shot performance.

\section{Experiments on MoEs}


In a MoE model, some or all of the feedforward networks (FFNs) in a standard monolithic model are replaced with MoE layers. 
As illustrated in \cref{fig:MoE}, a MoE layer is constructed by multiple experts and a gate, where each expert is essentially an FFN. 
The gate, typically implemented as a MLP, is responsible for selecting a specified number of experts to process the input. 
The outputs of the selected experts are then aggregated. 
The computation performed by a MoE layer can be expressed as \cite{shazeer2017outrageously}:
\begin{equation}
    y = \sum_{i=1}^n G(x)_iE_i(x)
\end{equation}
where $n$ denotes the total number of experts, $E_i(\cdot)$ represents the output of the $i$-th expert, and $G(\cdot)_i$ denotes the score computed by the gate for the $i$-th expert.

\subsection{Setups}

\textbf{Datasets.}
We exclusively used the ImageNet-1k dataset for MoE models throughout the pipeline (from preparation to E2E training). The few-shot adaptation ability is evaluated on both CIFAR-100 and CUB-2011 datasets \cite{wah2011caltech}.
To further examine the performance on downstream tasks, we fine-tune the models for COCO \cite{lin2014microsoft} object detection and instance segmentation.

\textbf{Preparation.}
We choose a DeiT-Large model with 24 layers and 304M parameters as the teacher model. 
Instead of training the meta model and the teacher model from scratch, we utilize the released checkpoints of Deep Incubation, which achieve an accuracy of 83.9\% on ImageNet-1k. 
The teacher model is evenly divided into $L=4$ sub-modules, each consisting of 6 layers.
The meta model is a four-layer DeiT model with the same embedded dimension as the teacher model.

\textbf{m2mKD.} In this series of experiments, the target model $S$ is set as a Vision MoE Base (V-MoE-B) model \cite{riquelme2021scaling}. 
Instead of using the carefully designed gate proposed by \cite{riquelme2021scaling}, we employ the earliest introduced gate in \cite{shazeer2017outrageously} which purely performs a top-$k$ ($k=2$) operation and no additional constraint or balancing loss is applied:
\begin{equation}
    Gate(x) = Softmax(TopK(W \cdot x)).
\end{equation}
Our V-MoE-B model is composed of 12 MoE layers (i.e., all feed-forward layers in a DeiT-B are changed to MoE layers) and each of them contains 8 experts.
Hence, there are a total of 483M parameters in the student model. 
Note that the student model size is larger than the teacher model size.
Each three layers of the V-MoE-B model are grouped as a student module, resulting in a total of 4 student modules.
In this case, we have $d_M=1024$ and $d_S=768$, and thus a pair of stitch layers accounts for about 1.6M parameters.
As a result, there are around 122.4M parameters for each student module and 77.6M parameters for each teacher module.
The two hybrid networks $\tilde{M}_S^{(i)}$ and $\tilde{M}_T^{(i)}$ constructed in this phase will then comprise 160.5M and 115.9M parameters respectively. 

Unlike the NAC models, the student modules of MoE do not require modification, and all of their learned parameters can be loaded into the target model during the end-to-end training phase.
The same as the settings for NAC models, we set the balancing factor $\alpha=0.5$ and softmax temperature $\tau=1.0$. 
All of the remaining hyperparameters are identical to those used in Deep Incubation for incubating DeiT-B \cite{touvron2021training} and the student modules are trained for 100 epochs. 

\textbf{E2E training.} 
Again, we adopt the same set of hyperparameters as Deep Incubation to train the V-MoE-B model, except for the update frequency argument, which is set to half of the original value.

\subsection{Results} \label{sec:moe-res}


\textbf{Main results.} We compare \texttt{m2mKD} with three baselines: pure end-to-end training, conventional knowledge distillation, and Deep Incubation.
For end-to-end training, we train a V-MoE-B model from scratch for 300 epochs using the same hyperparameters as in the DeiT-B training \cite{touvron2021training}.
In contrast to \texttt{m2mKD}, which performs knowledge distillation at the module level, conventional KD refers to knowledge distillation between complete models.
For the sake of fair comparison, we again use DeiT-L as the teacher model and V-MoE-B as the student model.
The KL divergence between their output logits is incorporated into the loss with $\alpha=0.5$ and $\tau=1.0$ (see \cref{Eq:loss}).
The training process lasts for 300 epochs. 
Lastly, we consider the original Deep Incubation approach, where we use their open source checkpoint of the meta model, originally trained for incubating DeiT-B, for the V-MoE-B experiments.
The validation results on ImageNet-1k are summarized in \cref{tab:moe-res}. 
It can be found that the V-MoE-B trained by the pure end-to-end method falls short of its monolithic counterpart, ViT-B, which achieves 81.8\% accuracy on ImageNet-1k \cite{touvron2021training}. 
This highlights the challenges faced during the training of modular models.
Conventional KD underperforms the other methods, with an accuracy 1.9\% lower than pure end-to-end training, which is the second lowest. 
This verifies our assertion that existing knowledge distillation techniques are not necessarily compatible with modular models.
On the other hand, the Deep Incubation approach achieves 2.97\% higher accuracy than end-to-end training, demonstrating its effectiveness not only for monolithic models, but also for modular models.
Our \texttt{m2mKD} approach further increases the accuracy by 0.50\%, resulting in the best performance among these methods. 
Given the remarkable OOD results of NACs in previous experiments, we also investigate the OOD robustness of MoEs on the ImageNet-R dataset, which is rare to be discussed in other MoE-related literatures. 
Surprisingly, all the MoE models trained using the four different methods achieve near 0\% accuracy on the ImageNet-R dataset.
Based on these experiments, it appears that NACs are significantly stronger than MoEs in terms of OOD or zero-shot performance. 

\begin{table}[tb]
    \centering
    \caption{Performance of V-MoE-B models trained using different approaches on ImageNet-1k. Baselines include pure end-to-end training (pure E2E), conventional knowledge distillation (KD), and Deep Incubation.}
    \label{tab:moe-res}
    \begin{tabular}{lccc}
        \toprule
         & Acc@1 & Acc@5 & Ratio of Time \\
        \midrule
        Pure E2E & 78.43 & 93.47 & 1.00 \\
        KD & 76.53 & 92.65 & 1.12 \\
        Incubation & 81.40 & 95.06 & 0.91 \\
        \texttt{m2mKD} & \textbf{81.90} & \textbf{95.43} & 0.99 \\
        \bottomrule
    \end{tabular}
\end{table}

\textbf{Training time.} In addition to accuracy, \cref{tab:moe-res} presents the training time ratios relative to the pure end-to-end training approach.
\texttt{m2mKD} is slightly slower than Deep Incubation but comparable to pure end-to-end training in terms of time, while conventional KD requires 1.12$\times$ time. 
Further experiments (Appendix E) demonstrate that the \texttt{m2mKD} ratio can be reduced to 0.57 with marginal performance degradation. 
For a fair comparison, we exclude the duration of teacher model training in the time calculation for KD and \texttt{m2mKD} methods. 
Hence, only the time for meta model training, distillation, and end-to-end training phases are considered for m2mKD.
It is worth questioning whether the preparation phase of \texttt{m2mKD} can be simplified. 
Currently, the \texttt{m2mKD} pipeline involves incubating a teacher model. 
We suppose such a model would be a more knowledgeable teacher for modular students. 
However, utilizing publicly available pretrained models as the teacher model instead of training a new one from scratch could save considerable time.
If this approach is adopted, the total training time would be exactly as stated in the table. 
We leave the investigation of how teachers trained using different methods influence \texttt{m2mKD} performance in future work.

\textbf{Few-shot.} 
We evaluate the few-shot adaptation of three models: a DeiT-B model trained using Deep Incubation (DeiT-DI), and two V-MoE-B models trained using Deep Incubation (MoE-DI) and \texttt{m2mKD} (MoE-Ours), respectively. 
The hyperparameters are listed in Appendix B.
During the few-shot training, all parameters in these models are frozen, except for the newly random initialized classifier layer, whose output dimension is set to the number of few-shot classes (8 in all experiments). 
The experiments for each shot are repeated with five different seeds. 
The average accuracy for each shot is depicted in \cref{fig:moe-fewshot}. 
Surprisingly, the DeiT-B model outperforms both V-MoE-B models, which are expected to be capable due to their modularity. 
This might be attributed to the relatively small dataset size (ImageNet-1k), resulting in insufficient training samples for each expert to learn comprehensively. 
However, when focusing on the two MoE models, MoE-Ours consistently outperforms MoE-DI by approximately 2\% for all shots on CUB-2011, as well as the 1-shot experiment on CIFAR-100. 
Similar to the few-shot results of NACs, the standard deviations are relatively large, ranging from 2.59 to 12.64. 
Conducting experiments on a larger dataset is necessary to obtain a more precise evaluation of the few-shot performance. 

\begin{figure*}[tb]
    \centering
    \includegraphics[width=.5\linewidth]{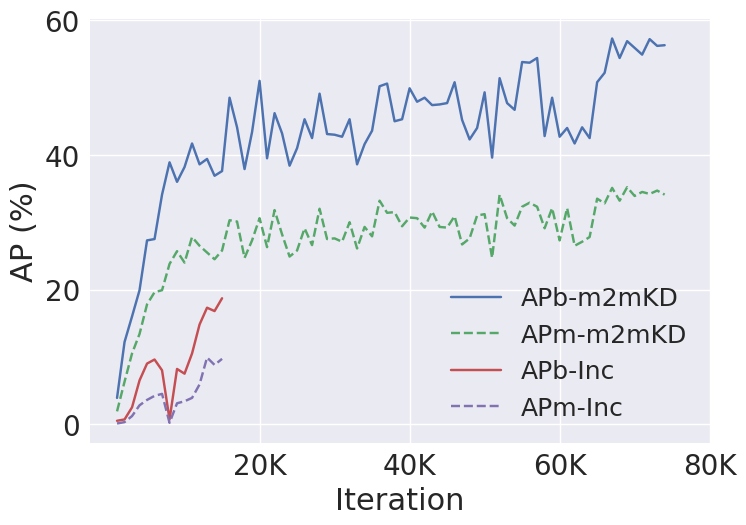}
    \caption{Validation accuracy of object detection (APb) and instance segmentation (APm) on COCO. Inc denotes Deep Incubation.}
    \label{fig:coco}
\end{figure*}

\textbf{Downstream tasks.}
We fine-tune the V-MoE-B models trained by Deep Incubation and \texttt{m2mKD} on the COCO dataset.
The training receipt is mostly the same as in \cite{li2022exploring}, except that we change the batch size from 64 to 8 since our available GPUs are limited.
Accordingly, the number of iterations are $8\times$ larger than the original to keep the total amount unchanged.
Figure~\ref{fig:coco} illustrates the validation accuracy across the whole fine-tuning process. 
The fine-tuning of Incubation-trained V-MoE-B is early terminated due to invalid gradients.
Nevertheless, the \texttt{m2mKD}-trained model consistently outperforms the Incubation model at the early stage and steadily gets stronger afterwards, suggesting the versatility of \texttt{m2mKD} beyond image classification.

\textbf{Ablations.} In the proposed \texttt{m2mKD} pipeline, stitch layers are used only during distillation and subsequently discarded to avoid modifying the student model.
Although they typically account for a small fraction of parameters in large modular models (e.g., 1\% for our V-MoE-B student), their removal may lead to loss of knowledge acquired from the teacher, potentially impacting performance. 
To investigate this aspect, we conduct experiments preserving stitch layers in the student model. 
However, the results indicate no noticeable improvement and thus we maintain our decision to discard them. 
Some additional experiments are carried out based on the setting of preserving stitch layers, including reducing distillation epochs and scaling up.
Decreasing the distillation epochs for \texttt{m2mKD} from 100 to 10 results in a 2.1\% accuracy drop, yet it remains 2.8\% higher than the Deep Incubation baseline.
To examine the scalability, we apply \texttt{m2mKD} to V-MoE-Large model and achieve 0.27\% higher accuracy on ImageNet-1k compared to Deep Incubation.
Given the negligible influence of preserving stitch layers, we believe that the above conclusions remain valid even if they are removed.
For further details on these ablation experiments, please refer to Appendix E. 

\section{Conclusion and Future Works}

In this work, we present module-to-module knowledge distillation (\texttt{m2mKD}) as a general approach for transferring knowledge in modular model training. 
\texttt{m2mKD} leverages a monolithic teacher model to facilitate the training of a modular student model, offering a promising strategy for transforming pretrained monolithic models into modular architectures to harness their advantages. 
Experimental results on NACs and MoEs demonstrate that the proposed pipeline enhances the IID accuracy and OOD robustness, even when the student model size exceeds that of the teacher model.
However, there is still room for narrowing the performance gap between the student and teacher. 
To further enhance effectiveness, combining \texttt{m2mKD} with techniques like ensemble learning holds potential.
We hope that this work can advance research on module-wise knowledge distillation and monolithic-to-modular conversion. 


\bibliographystyle{splncs04}
\bibliography{main}

\clearpage
\appendix

\section{Notations}

\begin{table}[htbp]
  \centering
  \caption{Notations commonly used in the main paper.}
  \label{tab:notations}
  \begin{tabular}{cl}
    \toprule
    Symbol & Definition \\
    \midrule
    $\mathcal{T}$ & Teacher model \\
    $T_i$ &  The $i$-th teacher module \\
    $\mathcal{M}$ & Meta model \\
    $M_i$ &  The $i$-th meta layer \\
    $\mathcal{S}$ & Student model \\
    $S_i$ &  The $i$-th student module \\
    $\tilde{S}_i$ & The $i$-th stitched student module \\
    $L$ & Number of student modules \\
    $d_M$ & Dimension of meta model which decides the size of stitch layers ($d_M \times d_S$) \\
    $d_S$ & Dimension of student model which decides the size of stitch layers ($d_M \times d_S$) \\
    $c_u$ & Code vector of the $u$-th module in a NACs model \\
    $s_u$ & Signature vector of the $u$-th module in a NACs model \\
    $\theta_u$ & State vector of the $u$-th module in a NACs model \\
    \bottomrule
  \end{tabular}
\end{table}

\newpage

\section{Training Hyperparameters}

\begin{table}[htbp]
  \centering
  \caption{A selection of hperparameters for the E2E training phase of NACs. When training on the same dataset, the hyperparameters remain unchanged for both the reproduced baselines and the distilled NAC models.}
  \label{tab:nacs-hyperparam}
  \begin{tabular}{lcc}
    \toprule
    Hyperparameter & Tiny-ImageNet & ImageNet \\
    \midrule
        Batch size & 1024 & 256 \\
        Number of epochs & 400 & 110 \\
    \midrule
        Optimizer & AdamW & AdamW \\ 
        Weight decay & 0.075 & 0.05 \\
        Learning rate scheduler & Cosine & Cosine \\
        Warmup epochs & 25 & 25 \\
        Warmup from learning rate &  1e-6 & 1e-6 \\
        Base peak learning rate & $0.0012 \times \frac{\text{batchsize}}{512}$ & $0.0012 \times \frac{\text{batchsize}}{512}$ \\
        Base min learning rate &  $4\mathrm{e}{-5} \times \frac{\text{batchsize}}{512}$ & $4\mathrm{e}{-5} \times \frac{\text{batchsize}}{512}$ \\
    \midrule
    Dimension of state & 384 & 512 \\
    Propagator layers & 8 & 8 \\
    Processor modules & 320 & 960 \\
    \quad - Input modules & 64 & 192 \\
    \quad - Propagator modules & 256 & 768 \\
    Attention heads & 6 & 8 \\
    Read-in heads & 6 & 8 \\
    Activation function & GEGLU & GEGLU \\ 
    FFN hidden units & 1536 & 1024 \\
    Output modules & 64 & 64 \\ 
    Signature dimension & 64 & 64 \\
    Code dimension& 384 & 512 \\
    Sampling temperature & 0.5 & 0.5 \\
    Kernel bandwidth & 1.0 & 1.0 \\
    Modulation at initialization & 0.1 & 0.1 \\ 
    \bottomrule
  \end{tabular}
\end{table}

\begin{table}[htbp]
  \centering
  \caption{Hyperparameters for the few-shot training of NAC, DeiT and V-MoE-B models.}
  \label{tab:fewshot-hyperparam}
  \begin{tabular}{lcc}
    \toprule
    & NAC & DeiT / V-MoE-B \\
    \midrule
    Batch size & $8 \times \text{shot}$ & $8 \times \text{shot}$ \\
    Number of epochs & 500 & 100 \\
    Optimizer & SGD & AdamW \\ 
    Weight decay & 0.0 & 0.05 \\
    Learning rate scheduler & None & Cosine \\
    (Peak) Learning rate & 0.0003 & 0.002 \\
    Min learning rate & N/A & $1\mathrm{e}{-5}$ \\
    Momentum & 0.9 & N/A \\
    Warmup epochs & N/A & 20 \\
    Warmup from learning rate & N/A & 1e-6 \\
    \bottomrule
  \end{tabular}
\end{table}

\newpage

\section{Reproducibility}

\begin{table}[htbp]
  \footnotesize
  \centering
  \caption{Mean and standard deviation of experimental results of NACs trained by \texttt{m2mKD} on Tiny-ImageNet. We repeat the end-to-end training phase of a distilled NAC model with a scale-free prior three times and calculate the mean and standard deviation of IID and OOD performance.}
  \label{tab:nacs-repeat-result}
  \begin{tabular}{lccccc}
    \toprule
    & \multicolumn{2}{l}{IID} & \multicolumn{2}{l}{OOD} \\
    & \multicolumn{2}{l}{(Tiny-ImageNet)} & \multicolumn{2}{l}{(Tiny-ImageNet-R)} \\
    \midrule
    & Acc@1 & Acc@5 & Acc@1 & Acc@5 \\
    \midrule
    1st & 66.47 & 85.08 & 24.31 & 44.38 \\
    2nd & 66.13 & 85.33 & 24.84 & 45.00 \\
    3rd & 65.94 & 85.49 & 25.29 & 45.61 \\
    \midrule
    MEAN & 66.18 & 85.30 & 24.81 & 45.00 \\
    STDEV & 0.27 & 0.21 & 0.49 & 0.62 \\
    \bottomrule
  \end{tabular}
\end{table}

\newpage

\section{Few-shot results}

\begin{table}[htbp]
    \centering
    \caption{Few-shot performance of NACs on CIFAR-100. The model is tasked with classifying the samples into 8 classes (i.e. 8-way).}
    \label{tab:nacs-fewshot-res}
    \begin{tabular}{ccccccccc}
        \toprule
        \multirow{2}{*}{Seed} & \multicolumn{2}{c}{1-shot} & \multicolumn{2}{c}{2-shot} & \multicolumn{2}{c}{4-shot} & \multicolumn{2}{c}{8-shot} \\
        & E2E & \texttt{m2mKD} & E2E & \texttt{m2mKD} & E2E & \texttt{m2mKD} & E2E & \texttt{m2mKD} \\
        \midrule
        1 & 53.66 & 60.82 & 64.39 & 64.39 & 71.21 & 75.06 & 81.03 & 80.53 \\
        2 & 60.82 & 50.08 & 64.39 & 59.03 & 75.06 & 78.91 & 84.04 & 82.03 \\
        3 & 67.97 & 67.97 & 75.13 & 67.97 & 78.91 & 79.87 & 80.53 & 79.03 \\
        4 & 60.82 & 60.82 & 67.97 & 57.24 & 75.06 & 78.91 & 84.04 & 82.53 \\
        5 & 50.08 & 60.82 & 60.82 & 60.82 & 73.14 & 78.91 & 80.03 & 79.53 \\
        \midrule
        MEAN & 58.67 & 60.10 & 66.54 & 61.89 & 74.68 & 78.33 & 81.93 & 80.73 \\
        STDEV & 6.98 & 6.40 & 5.43 & 4.31 & 2.85 & 1.88 & 1.95 & 1.52 \\
        \bottomrule
    \end{tabular}
\end{table}

\begin{table}[htbp]
    \centering
    \caption{Differences between two runs of the 2-shot experiment for NAC models.}
    \label{tab:nacs-fewshot-repeat}
    \begin{tabular}{ccccccc}
        \toprule
        \multirow{2}{*}{seed} & \multicolumn{3}{c}{E2E}& \multicolumn{3}{c}{\texttt{m2mKD}} \\
        & 1st & 2nd & $|\Delta|$ & 1st & 2nd & $|\Delta|$ \\
        \midrule
        1 & 64.39 & 69.76 & 5.37 & 64.39 & 66.18 & 1.79 \\
        2 & 64.39 & 64.39 & 0.00 & 59.03 & 60.82 & 1.79 \\
        3 & 75.13 & 75.13 & 0.00 & 67.97 & 75.13 & 7.16 \\
        4 & 67.97 & 66.18 & 1.79 & 57.24 & 62.60 & 5.36 \\
        5 & 60.82 & 62.60 & 1.78 & 60.82 & 64.39 & 3.57 \\
        \midrule
        MEAN & 66.54 & 67.61 & 1.07 & 61.89 & 65.82 & 3.93 \\
        STDEV & 5.43 & 4.97 & 0.46 & 4.31 & 5.57 & 1.26 \\
        \bottomrule
    \end{tabular}
\end{table}

\begin{table}[htbp]
    \scriptsize
    \centering
    \caption{Few-shot performance of DeiT and V-MoE-B models on CIFAR-100 (top) and CUB-2011 (bottom). The models are tasked with classifying samples into 8 classes (i.e., 8-way). We compare three models: a DeiT-B model trained using Deep Incubation (DeiT-DI), and two V-MoE-B models trained using Deep Incubation (MoE-DI) and \texttt{m2mKD} (MoE-Ours), respectively.}
    \label{tab:moe-fewshot-res}
    \begin{tabular}{llllllllll}
        \toprule
        \multicolumn{10}{c}{CIFAR-100} \\
        \midrule
        \multirow{2}{*}{Seed} & \multicolumn{3}{l}{1-shot} & \multicolumn{3}{l}{5-shot} & \multicolumn{3}{l}{10-shot} \\
        & DeiT-DI & MoE-DI & MoE-Ours & DeiT-DI & MoE-DI & MoE-Ours & DeiT-DI & MoE-DI & MoE-Ours \\
        \midrule
        1 & 70.83 & 70.83 & 70.83 & 90.83 & 90.00 & 89.17 & 90.83 & 91.67 & 90.42 \\
        2 & 70.83 & 62.50 & 66.37 & 90.00 & 87.50 & 86.67 & 92.92 & 92.50 & 90.42 \\
        3 & 87.50 & 95.83 & 91.67 & 95.83 & 94.17 & 94.17 & 97.08 & 96.25 & 96.67 \\
        4 & 95.83 & 79.17 & 87.50 & 95.83 & 94.17 & 94.17 & 97.50 & 94.58 & 96.25 \\
        5 & 79.17 & 70.83 & 70.83 & 88.33 & 87.50 & 87.50 & 90.00 & 89.58 & 90.42 \\
        \midrule
        MEAN & 80.83 & 75.83 & 77.44 & 92.16 & 90.67 & 90.34 & 93.67 & 92.92 & 92.84 \\
        STDEV & 10.87 & 12.64 & 11.33 & 3.47 & 3.36 & 3.61 & 3.48 & 2.59 & 3.31 \\
        \bottomrule
    \end{tabular}
    \begin{tabular}{llllllllll}
        \toprule
        \multicolumn{10}{c}{CUB-2011} \\
        \midrule
        \multirow{2}{*}{Seed} & \multicolumn{3}{l}{1-shot} & \multicolumn{3}{l}{5-shot} & \multicolumn{3}{l}{10-shot} \\
        & DeiT-DI & MoE-DI & MoE-Ours & DeiT-DI & MoE-DI & MoE-Ours & DeiT-DI & MoE-DI & MoE-Ours \\
        \midrule
        1 & 83.33 & 70.83 & 70.83 & 90.83 & 90.83 & 91.67 & 95.00 & 94.38 & 93.75 \\
        2 & 79.17 & 83.33 & 87.5 & 94.17 & 93.33 & 94.17 & 98.75 & 95.63 & 98.13 \\
        3 & 95.83 & 91.67 & 95.83 & 95.00 & 92.50 & 92.50 & 94.38 & 91.25 & 91.25 \\
        4 & 87.50 & 75.00 & 79.17 & 88.33 & 80.83 & 84.17 & 84.38 & 80.63 & 86.25 \\
        5 & 66.67 & 70.83 & 75.00 & 90.83 & 88.33 & 94.17 & 96.25 & 93.75 & 95.63 \\
        \midrule
        MEAN & 82.50 & 78.33 & 81.67 & 91.83 & 89.16 & 91.34 & 93.75 & 91.13 & 93.00 \\
        STDEV & 10.78 & 9.04 & 10.03 & 2.73 & 5.04 & 4.15 & 5.50 & 6.08 & 4.54 \\
        \bottomrule
    \end{tabular}
\end{table}

\newpage

\section{Ablation Studies} \label{append:ablation}

In the proposed \texttt{m2mKD} pipeline, stitch layers are used only during distillation and subsequently discarded to avoid modifying the student model.
Although they typically account for a small fraction of parameters in large modular models (e.g., 1\% for our V-MoE-B student), their removal may lead to loss of knowledge acquired from the teacher, potentially impacting performance. 
To investigate this aspect, we conduct experiments preserving stitch layers in the student model. 
However, as shown in \cref{tab:ablation}, the results indicate no noticeable improvement from this approach. 
Consequently, we maintain our decision to discard the stitch layers. 

The remaining ablation experiments are conducted based on the preservation of stitch layers. 
Given the negligible influence of preserving stitch layers, we believe that the following conclusions remain valid even if they are removed.
Firstly, since the NAC models demonstrate considerable improvement with student modules distilled for only 10 epochs, we test the performance of V-MoE-B models under the same condition. 
To this end, we adjust some of the hyperparameters during the \texttt{m2mKD} phase and distill the student module for 10 epochs, while retaining the E2E phase at 100 epochs. 
As shown in \cref{tab:ablation}, decreasing the distillation epochs from 100 to 10 results in a drop of 2.1\% in accuracy, yet it remains 2.8\% higher than the Deep Incubation baseline. 
Surprisingly, applying the same changes to the incubation phase of Deep Incubation does not lead to any performance degradation. 
Next, we examine the scalability of \texttt{m2mKD}. 
The teacher model is the DeiT-Huge model (632M) pretrained by Deep Incubation, while the student model ($L=4$) is a V-MoE-Large model with 12 MoE layers placed on every other layer, resulting in a total of 1.0B parameters. 
As presented in \cref{tab:ablation}, \texttt{m2mKD} outperforms Deep Incubation by 0.27\% in terms of accuracy on ImageNet-1k.

\begin{table}[htbp]
    \centering
    \caption{Results of ablation studies. The hyperparameters adjusted for decreasing distillation epochs are as follows: warm-up-epochs=5, warm-up-lr=0.0001, min-lr=0.002. The ratio of training time is calculated relative to the time used in the pure end-to-end training approach. The column for V-MoE-L is empty because we did not train our V-MoE-L model using the pure end-to-end method. Note that all experiments, except the \textit{baseline}, are conducted with stitch layers inserted in the student model.}
    \label{tab:ablation}
    \begin{tabular}{llcccc}
        \toprule
         & & Acc@1 & Acc@5 & Extra params & Ratio of Time \\
        \midrule
        \multirow{3}{*}{\texttt{m2mKD}} & baseline & 81.90 & 95.43 & 0 & 0.99 \\
        & w/ stitch &  81.93 & 95.53 & 4.7M (1\%) & 0.99 \\
        & 10 epochs & 81.72 & 95.44 & 0 & 0.57 \\
        \midrule
        \multirow{2}{*}{Deep Incubation} & baseline & 81.40 & 95.06 & 0 & 0.91 \\
        & 10 epochs & 81.44 & 95.09 & 0 & 0.54 \\
        \midrule
        \multirow{2}{*}{V-MoE-L} & \texttt{m2mKD} & 83.36 & 96.42 & 7.9M (0.8\%) & N/A \\
         & Deep Incubation & 83.09 & 96.12 & 0 & N/A \\
        \bottomrule
    \end{tabular}
\end{table}

\end{document}